%% file: main.tex
\documentclass{article}

     \usepackage[final,nonatbib]{neurips_2019}

\usepackage[utf8]{inputenc} 
\usepackage[T1]{fontenc}    
\usepackage{hyperref}       
\usepackage{url}            
\usepackage{booktabs}       
\usepackage{amsfonts}       
\usepackage{nicefrac}       
\usepackage{microtype}      
\usepackage{graphicx}

\usepackage{times}
\usepackage{amsmath}
\usepackage{amssymb}

\usepackage{listings}
\usepackage{xcolor}

\colorlet{punct}{red!60!black}
\definecolor{background}{HTML}{EEEEEE}
\definecolor{delim}{RGB}{20,105,176}
\colorlet{numb}{magenta!60!black}

\lstdefinelanguage{json}{
    basicstyle=\normalfont\ttfamily,
    numberstyle=\scriptsize,
    stepnumber=1,
    numbersep=8pt,
    showstringspaces=false,
    breaklines=true,
    frame=lines,
    backgroundcolor=\color{background},
    literate=
     *{0}{{{\color{numb}0}}}{1}
      {1}{{{\color{numb}1}}}{1}
      {2}{{{\color{numb}2}}}{1}
      {3}{{{\color{numb}3}}}{1}
      {4}{{{\color{numb}4}}}{1}
      {5}{{{\color{numb}5}}}{1}
      {6}{{{\color{numb}6}}}{1}
      {7}{{{\color{numb}7}}}{1}
      {8}{{{\color{numb}8}}}{1}
      {9}{{{\color{numb}9}}}{1}
      {:}{{{\color{punct}{:}}}}{1}
      {,}{{{\color{punct}{,}}}}{1}
      {\{}{{{\color{delim}{\{}}}}{1}
      {\}}{{{\color{delim}{\}}}}}{1}
      {[}{{{\color{delim}{[}}}}{1}
      {]}{{{\color{delim}{]}}}}{1},
}

\title{AuthorGAN: Improving GAN Reproducibility using a Modular GAN Framework}

\author{%
  Raunak Sinha, Anush Sankaran\\
  IBM Research AI\\
  Bengaluru, India\\
  \texttt{\{rsinha05, anussank\}@in.ibm.com} \\
  \And
  Mayank Vatsa, Richa Singh \\
  IIIT Delhi \\
  New Delhi, India \\
  \texttt{\{mayank, rsingh\}@iiitd.ac.in} \\
}

\begin{document}

\maketitle

\begin{abstract}
 Generative models are becoming increasingly popular in the literature, with Generative Adversarial Networks (GAN) being the most successful variant, yet. With this increasing demand and popularity, it is becoming equally difficult and challenging to implement and consume GAN models. A qualitative user survey conducted across 47 practitioners show that expert level skill is required to use GAN model for a given task, despite the presence of various open source libraries. In this research, we propose a novel system called AuthorGAN, aiming to achieve true democratization of GAN authoring. A highly modularized library agnostic representation of GAN model is defined to enable interoperability of GAN architecture across different libraries such as Keras, Tensorflow, and PyTorch. An intuitive drag-and-drop based visual designer is built using node-red platform to enable custom architecture designing without the need for writing any code. Five different GAN models are implemented as a part of this framework and the performance of the different GAN models are shown using the benchmark MNIST dataset.
\end{abstract}

\graphicspath{{./images/}}
\input{introduction}
\input{user_survey}
\input{architecture}

\input{visual}

\input{gan_models}
\input{performance}
\input{background}
\input{conclusion}

\bibliographystyle{acm}
\bibliography{gan_models}

\end{document}

%% file: introduction.tex
\section{Introduction}

Automated generative models have progressed a lot over the last couple of years~\cite{goodfellow2016nips}. Many variations of generative models have been proposed and discussed in the literature such as Gaussian Mixture Models (GMM)~\cite{reynolds2015gaussian}, Hidden Markov model (HMM)~\cite{HMM1,HMM2}, Latent Dirichlet Allocation (LDA)~\cite{blei2003latent}, Restricted Boltzmann Machine (RBM)~\cite{RBM1,RBM2}. A recent successful formulation of generative models is called as Generative Adversarial Networks (GAN). GAN models contains a generative module and a discriminative module competing with each other. When equilibrium between these two learning modules is achieved, the generator would have approximated the distribution of the given data. 

The initial research about GAN was proposed in 2014~\cite{VanillaGAN} and has rapidly grew in sophistication and complexity ever since, with more than $100$ named GANs in the literature. Currently, research works such as Progressive GAN~\cite{Progressive_GAN} can generate high resolution and almost-real looking images from nothing but just random numbers as input.
With the advent of so many GAN models, reproducibility and easy consumability becomes a challenge. 
\textit{``How easy is it to use a known GAN model for my task?"} and \textit{``What are the coding and technical prerequisites required to work on GAN?"} are the two of the many open challenges for using GAN models.
Despite the presence of GAN zoo's such as TFGAN~\footnote{\url{https://github.com/tensorflow/tensorflow/tree/master/tensorflow/contrib/gan}}, Keras-GAN~\footnote{\url{https://github.com/eriklindernoren/Keras-GAN}}, PyTorch-GAN\footnote{\url{https://github.com/eriklindernoren/PyTorch-GAN}}, building custom GAN models is challenging for the following reasons: (i) \textbf{to implement from scratch,} there is a requirement for expert level understanding of the libraries such as PyTorch, Keras, or Tensorflow, (ii) \textbf{modifying an existing GAN}, is challenging as the entire GAN model is tightly coupled at a code-level and modifying a small component such as only generator, or discriminator, or a loss function requires tinkering of the entire code, (iii) \textbf{mix-and-match} of different GAN components is not possible as there is no universal GAN architecture and also models implemented in different libraries are not interoperable with each other. Due to these challenges, there is huge learning curve for software engineers to start implementing and playing around with GAN models. Also, researchers struggle in reproducing and comparing results of different GAN models for their research papers.

Authoring GAN models is a struggling and time taking experience for both novice and expert users. There is a need for a easy-to-use authoring system that is agnostic of library or language underneath and gives a very intuitive interface for users to play around with GAN models. Thus, in this paper we propose, \textit{AuthorGAN}\footnote{The \textit{AuthorGAN} system will be made available as an open source system upon the acceptance of this paper.}, a GAN authoring system with the following research contributions:
\begin{enumerate}
    \item To modularize the GAN architecture into reusable components and propose a library-agnostic, abstract, modularized representation for GAN models,
    \item Develop a highly extensible, no-code system implementing GAN models. The toolkit has an intuitive visual interface for authoring GAN models from scratch without the need for writing code in any library. The toolkit is available here: \url{}
    \item A qualitative user survey detailing the challenges faced in developing and implementing GAN models by different kinds of users,
    \item Experimental results to quantify the performance of the different GAN models developed using our system on benchmark MNIST and dataset.
\end{enumerate}



%% file: user_survey.tex
\section{Preliminary User Study}

\begin{figure}[!t]
	\begin{center}
	\includegraphics[width=.9\textwidth]{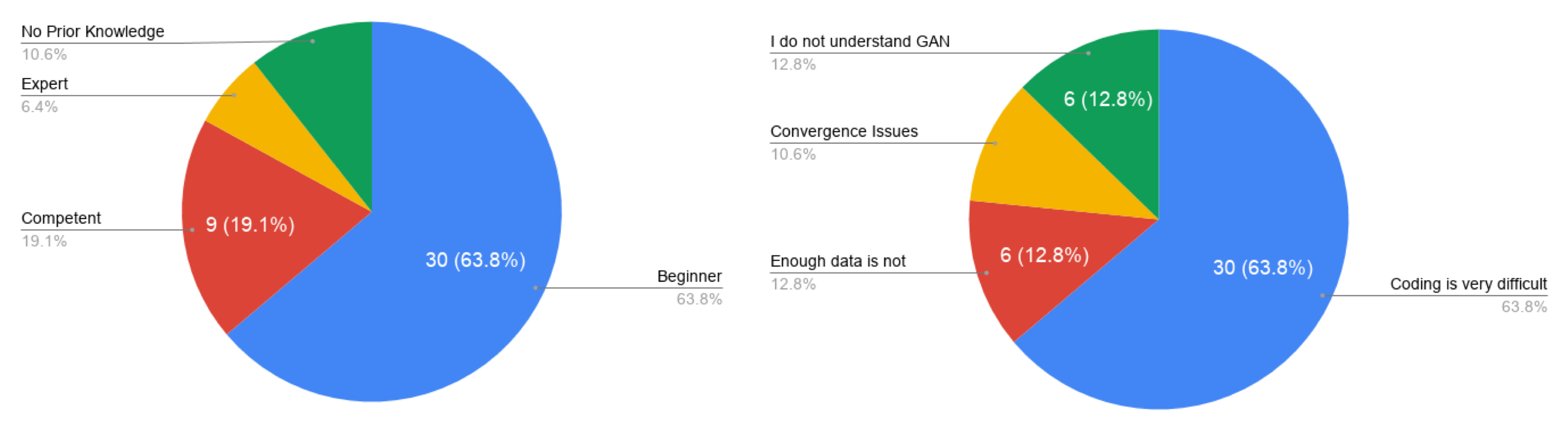}
	\end{center}
	\caption{(On the left) The distribution of the participants based on their expertise with GAN models in the qualitative user survey conducted. (On the right) The different challenged faced by the user survey participants in authoring and consuming GAN models.}
	\label{fig:chart1}
\end{figure}

The initial step is to understand the challenges involved in developing and consuming GAN models from the different kinds of users. A quantitative survey was conducted across $47$ different developers, software engineers, and researchers from various organizations and academic institutions. $76\%$ of the participants were male with a very diverse age distribution between $20$-$38$. The survey was conducted among those participants who had minimum knowledge about GANs with almost $25\%$ of them rating themselves as experts, as shown in Figure~\ref{fig:chart1}, and $50\%$ having hands-on experience with GANs. It was quite interesting to note that almost $30\%$ (highest) learnt about GANs from research papers while only $14\%$ learnt it from basic course material and tutorials. In fact, as a free text feedback many participants requested for the need of a MOOC or a structured course in learning GANs.


An overwhelming $72\%$ of the participants expressed that they had challenges. As shown in Figure~\ref{fig:chart1}, out of the various challenges expressed, the most popular one was ``Coding is very difficult" ($64\%$) while author challenges included ``Enough data is not available" ($13\%$) and ``I do not understand GAN" ($13\%$). This motivates the need for a no-coding based system for developing GAN models as developers and researchers find it non-intuitive to code GAN models with minimal learning. Additionally, the system should intuitive to use with detailed documentation for consumers who find it difficult to understand GANs. It is interesting to note that $73\%$ of the participants who rated themselves as ``competent" or ``expert" voted that it was not easy for them to code GAN model or that they eventually gave up. Some of the free text feedback that the participants provided as shown here,
\begin{itemize}
    \item "An easy to use interface in which we can specify model parameters and architecture quickly without much coding."
    \item "Easy to use Interface to stitch components of GAN's with easy to edit logic/parameters"
    \item "Easily and quickly change the architecture and hyperparameters to experiment models."
    \item "Given dimensionality/ modality of the input and the output, if the system could predict the configuration of GAN to be used, it would be really great!"
\end{itemize}

Thus, there is a need for having an intuitive system that could democratize the development of GAN authoring and make GAN consumable by users with different technical prowess.

%% file: architecture.tex

\begin{figure*}[!t]
	\begin{center}
 	\includegraphics[width=.9\textwidth]{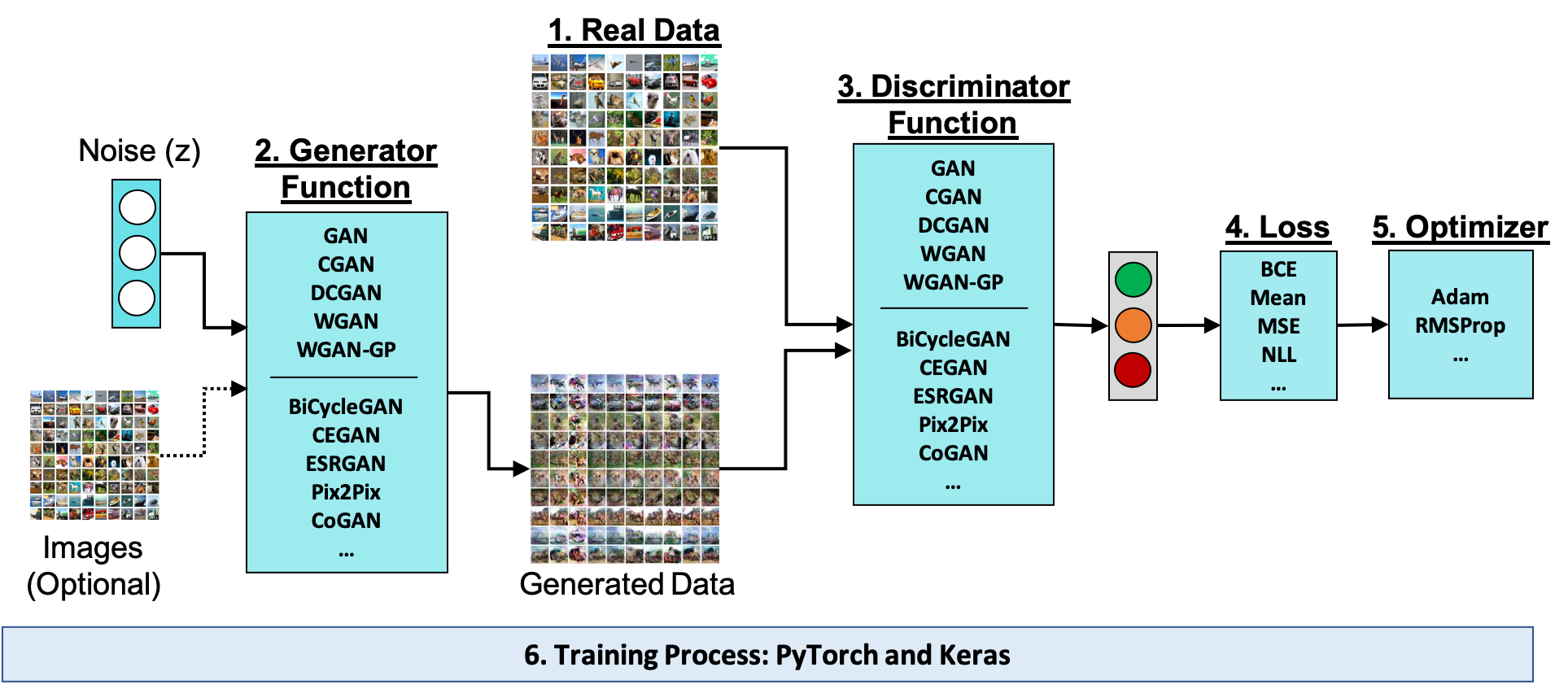}
 	\end{center}
 	\caption{The different mix-and-match GAN models that could be implemented using our modular GAN framework.}
 	\label{fig:dcgan}
\end{figure*}

\section{System Architecture}


\subsection{Modularization of GAN Architecture}

After studying different named GAN variations, we identified that a standard GAN model would contains six different modules, as follows:

\begin{enumerate}
    \item \textbf{Real training data:} This is the input data whose distribution is to be learnt and regenerated. Some examples include MNIST dataset, CIFAR-10 dataset, or movie review text dataset.
    \item \textbf{Generator:} It is a generative function that takes a random noise vector (latent space) as input and generates an output that is of the same modality and dimensions of the real training data. Some examples include Deconvolutional network, or recurrent neural network.
    \item \textbf{Discriminator:} This is typically a classifier which learns from the real training data and the data generated by the generator (fake data). The primary task of the discriminator is distinguish between the real data and fake data. Some examples include multi-layer perceptron, convolutional neural network, or regression.
    \item \textbf{Loss function:} Two different loss functions are used; one for computing the generator's loss and the other for the discriminator's loss. Some standard loss functions include euclidean loss, cross-entropy loss, or hinge loss.
    \item \textbf{Optimizer function:} Two different optimizers are used to learn the generator and discriminator, seperately. Some standard optimizers are SGD or RMSProp.
    \item \textbf{Training process:} If both the generator and the discriminator adopt a network based architecture, standard back-propogation based techniques could be used for a model update. However, certain GANs required much sophisticated training process such as reinforcement learning. The overall learning process and its parameters are detailed in this module. 
\end{enumerate}

The primary idea of GAN modularization is that the different modules could be combined seamlessly and a mix-and-match GAN architecture could be defined easily. For example, the generator of the DCGAN can be combined with the discriminator of a WGAN with the loss functions and optimizers from a CGAN to build a novel GAN architecture.

\subsection{Library Agnostic Abstract GAN Representation}

Consider the example of a popular Deep Convolutional GAN (DCGAN) model, as shown in Figure~\ref{fig:dcgan}. The PyTorch implementation of the model would roughly contain $150$ lines of code\footnote{\url{https://github.com/pytorch/examples/blob/master/dcgan/main.py}} and the tensorflow implementation would require $500$ lines of code\footnote{\url{https://github.com/carpedm20/DCGAN-tensorflow/blob/master/model.py}}, with the requirement of expertise for model customization. However, we propose a simple JSON representation of defining a GAN model, extending the modules explained in the previous section. The most simplistic realization of the DCGAN architecture is shown below:

 \begin{lstlisting}[language=json]
{
    "GAN_model":{
        "epochs":"50"
    },
    "generator":{
        "choice":"dcgan"
    },
    "discriminator":{
        "choice":"dcgan"
    },
    "data_path":"dataset/mnistData.pkl"
}
\end{lstlisting}

Providing this config file as the only input to our AuthorGAN system is enough to train a GAN model and obtain the performance results. There is no need to write even a single line of code. Thus, defining this JSON object does not require any expertise in Python, PyTorch, or GANs. Additionally, this JSON representation is library agnostic and as shown in Figure~\ref{fig:dcgan}, multiple library drivers (Keras/ PyTorch) could be written to parse the JSON object into the respective static computational graphs. Moreover, the proposed abstract representation provides enough flexibility to define every configurable parameter of the DCGAN model. Thus, the abstract representation not only provides the easiness of authoring GAN model for novice level users but also offers flexibility to provide detailed parameters for expert level users.


%% file: visual.tex
\section{Visual Authoring}

\begin{figure*}[!t]
	\begin{center}
	\includegraphics[width=.9\textwidth]{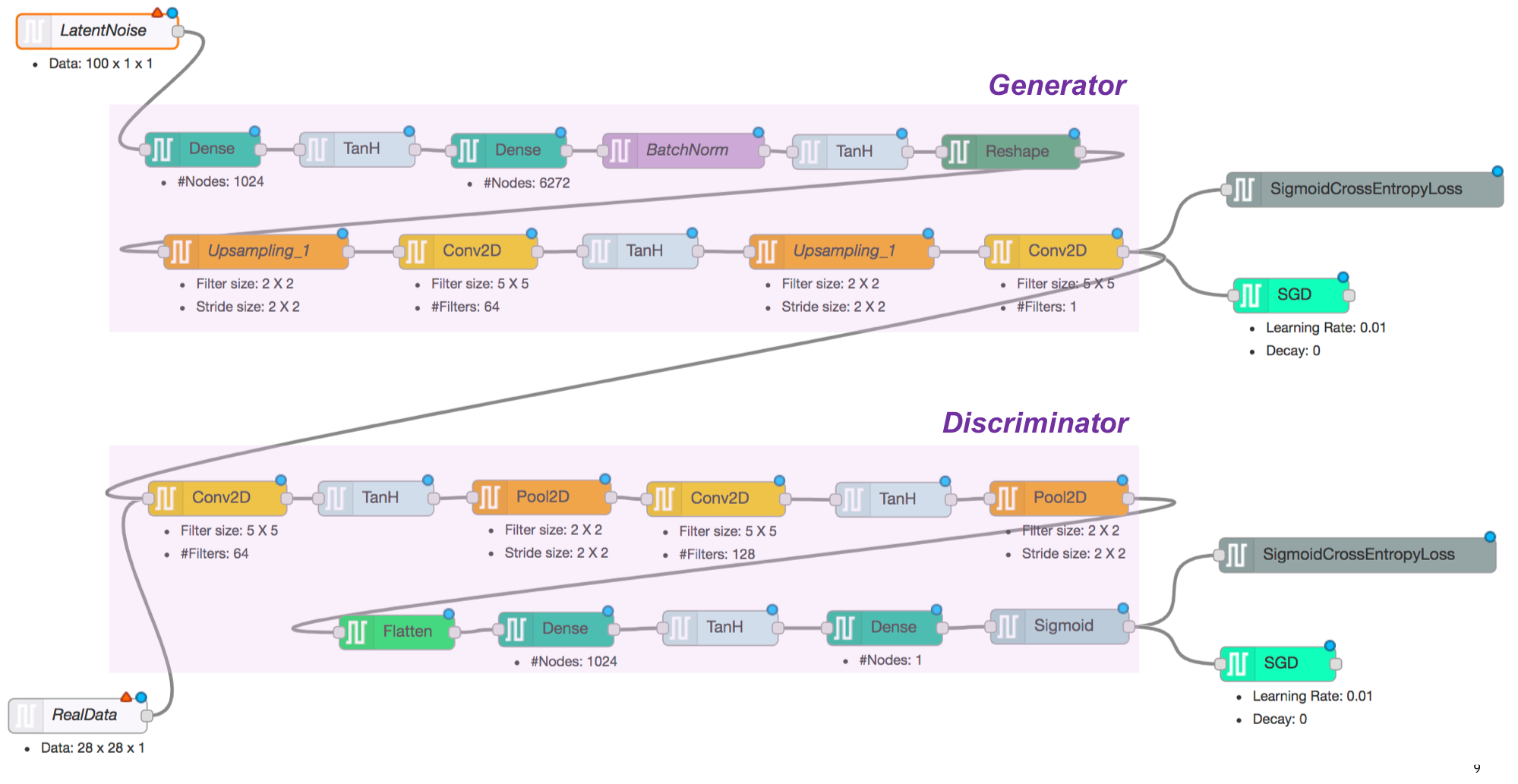}
	\end{center}
	\caption{An example illustrating the power of the visual authoring capability of AuthorGAN to design the DCGAN model~\cite{DCGAN} in an from scratch.}
	\label{fig:design2}
\end{figure*}

The JSON object based GAN model representation offers abstraction over multiple libraries and enables cross library interoperability. However, true democratization of GAN authoring requires an intuitive authoring interface for entry level users to adopt GAN models. Thus, as shown in Figure~\ref{fig:design2}, we developed an drag-and-drop interface to author GAN models. The drag-and-drop interface is built over a powerful open source node-red platform\footnote{\url{https://nodered.org/}}. All the required functionality are available as a palette of nodes, from which the users could chose and design an architecture in an intuitive fashion. Each node (function) could be parameterized through the user interface, avoiding the task of creating the abstract JSON object from scratch. The node-red based user interface internally generates the JSON object which is used by the backend library drivers for model creation and training. 

Further, customizing GAN models and authoring GAN models from scratch is super fun and easy using the user interface of AuthorGAN. As shown in Figure~\ref{fig:design2}, the basic layers of a neural network architecture are provided in the node palette. A list of $31$ layers is provided, grouped under seven categories: (i) convolutional layers, (ii) recurrent layers, (iii) core layers, 
(iv) activation layers, (v) loss layers, (vi) optimization layers, and (vii) normalization layers. GAN architectures could be designed from scratch and the backend drivers could additionally read the custom generator or discriminator architecture and create the static computational graphs in the respective libraries. Thus, GAN models could be visually authored using the intuitive user interface of AuthorGAN system. To the best of our knowledge, this is the first visual authoring system for designing and building GAN models.

%% file: performance.tex
\section{Experimental Analysis}

The performance of the different implemented GAN models is shown using the benchmark MNIST handwritten image classification dataset. The task of each GAN would be to generate digit images similar to the MNIST dataset. We implement five different GAN models using our AuthorGAN frameowork: Vanilla GAN (GAN) \cite{VanillaGAN}, Conditional GAN (CGAN)~\cite{CGAN}, Deep Convolutional GAN (DCGAN)~\cite{DCGAN}, Wasserstein GAN (WGAN)~\cite{WGAN}, Wasserstein GAN with Gradient Policy (WGAN-GP)~\cite{WGAN_GP}. Addtionally, to show the flexbility of the proposed modularized AuthorGAN system, different mix-and-match GAN models are formed and their experimental performance is shown in this section.

A total of $16$ different GAN configurations are created and is defined as $<$generator, discriminator$>$: 1. GAN, GAN; 2. GAN, DCGAN; 3. GAN, WGAN; 4. GAN, WGAN\_GP; 5. WGAN, GAN; 6. WGAN, DCGAN; 7. WGAN, WGAN; 8. WGAN, WGAN\_GP; 9. WGAN\_GP, GAN; 10. WGAN\_GP, DCGAN; 11. WGAN\_GP, WGAN; 12. WGAN\_GP, WGAN\_GP; 13. DCGAN, GAN; 14. DCGAN, DCGAN; 15. DCGAN, WGAN; 16. DCGAN, WGAN\_GP.


\begin{figure}[!t]
	\begin{center}
	\includegraphics[width=0.95\textwidth]{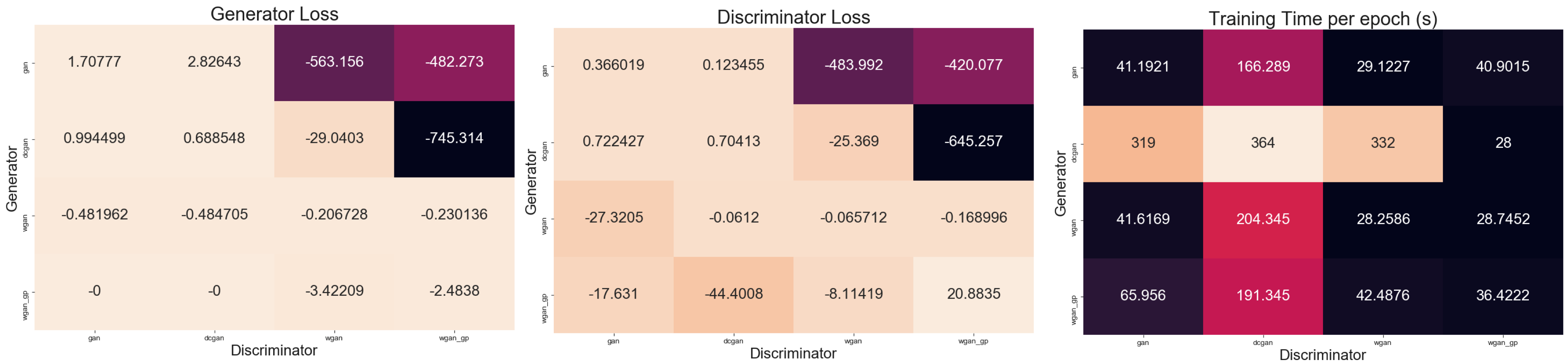}
	\end{center}
	\caption{(On the left) Generator loss, (On the middle) Discriminator loss, (On the right) Average training time per epoch in seconds; obtained for different GAN combinations implemented using the proposed system.}
	\label{fig:exp2}
\end{figure}

To compare the performance of the different configurations, a heatmap confusion matrix of the final generator and discriminator loss is shown in Figure~\ref{fig:exp2}. Using other GAN libraries, only the diagonal elements of this confusion matrix could be obtained while obtaining the other elements of the confusion matrix is not straight forward. This demonstrates both the performance as well as the flexibility of the AuthorGAN system.

In terms of training time, the average time required to train one epoch (in seconds) is shown Figure~\ref{fig:exp2}. It can be observed that the configurations that use DCGAN's generator or discriminator train much slower than the other configurations. This benchmark helps us in comparing different GAN model not only in terms of performance accuracy but also in terms of efficiency.

%% file: background.tex
\section{Discussion}

As a testimonial to the importance of this problem and the need for easy-to-use systems there are a few GAN libraries in the literature, such as, PyTorchGAN, TFGAN, and KerasGAN. These open source libraries provide a collection of existing GAN implementation in their respective libraries. However as compared to the existing libraries, the proposed AuthorGAN system provides the following advantages,
\begin{enumerate}
    \item Highly modularized representation of GAN model for easy mix-and-match of components across architectures. For instance, one can use the generator component from DCGAN and the discriminator component from CGAN, with the training process of WGAN. While this could be done in other libraries, typically, they are coding intensive and require expert level knowledge.
    \item An abstract representation of GAN architecture to provide multi-library support. Currently, we are providing backend PyTorch and Keras support for the provided JSON object, while in the future we plan to support Tensorflow, Caffe and other libraries as well. Thus, the abstract representation is library agnostic.
    \item Coding free, visual designing of GAN models. A highly intuitive drag-and-drop based visual designer is provided to author GANs and there is no need for writing any code to train the GAN model.
\end{enumerate}

During the process of building this authoring system, there were a few interesting observations that we made and are discussed as follows:
\begin{enumerate}
    \item The quantitative user survey conducted, additionally points out that using evaluation metrics to evaluate GAN models and using transfer learning to speed up the training process are not popularly known among GAN users. Thus, it is required to focus on these aspects of the system in the successive versions.
    \item The participants of the user survey showed equal interest in using three different libraries: PyTorch, Keras, and Tensorflow. This represents that there is a requirement for a stronger knitting between these three libraries and the proposed AuthorGAN system could act as the entry point to use any of these library models.
    \item It is to be understood that this is a continuously evolving system. The usability could be achieved by adding more state-of-art GAN models into this system. Thus, we plan to make this whole system and framework open source, to benefit the entire GAN community.
\end{enumerate}

%% file: conclusion.tex
\section{Conclusion and Future Directions}

Thus, in this research an AuthorGAN system is proposed and developed, to achieve true democratization of authoring GAN models. A highly modular and abstract GAN represented was defined to allow interoperability across different dimensions. An easy-to-use intuitive visual designer was developed to allow novice users construct custom GAN architectures. The benchmark results of various GANm models implemented using our system was shown on benchmark MNIST dataset. 

As immediate extension to this system, the following future directions are identified: (i) Implement atleast $50$ GAN models as a part of this system, by modularizing every GAN architecture, (ii) extend the framework to support different training process such as reinforcement learning, and (iii) support other data types such as text data and speech data to increase the scope of usage.